\documentclass[conference]{IEEEtran}

\IEEEoverridecommandlockouts                              

\usepackage[letterpaper, left=54pt, right=54pt, bottom=54pt, top=54pt]{geometry}

\usepackage{cite}
\usepackage{amsmath,amssymb,amsfonts}
\usepackage{algorithmic}
\usepackage{graphicx}
\usepackage{textcomp}
\usepackage{xcolor}
\usepackage{array}
\usepackage{tablefootnote}
\usepackage{caption}
\usepackage{booktabs}
\usepackage{arydshln}
\usepackage{url}
\usepackage{hyperref}
\newcommand{\etal}{\textit{et al}. }

\usepackage{tablefootnote}

\def\BibTeX{{\rm B\kern-.05em{\sc i\kern-.025em b}\kern-.08em
    T\kern-.1667em\lower.7ex\hbox{E}\kern-.125emX}}
\begin{document}

\title{\vspace*{18pt}Evaluation of an Inflated Beam Model\\ Applied to Everted Tubes
}

\author{\IEEEauthorblockN{
    Joel Hwee, 
    Andrew Lewis, 
    Allison Raines\textsuperscript{1}, 
    Blake Hannaford\textsuperscript{1}}
\thanks{Mechanical Engineering, \textsuperscript{1}Electrical and Computer Engineering}
\thanks{\textit{University of Washington}, Seattle, WA 98195 USA}
\thanks{(hweejo3, alewi, arain98, blake) (at) uw.edu}
}

\maketitle

\begin{abstract}
Everted tubes have often been modeled as inflated beams to determine transverse and axial buckling conditions. This paper seeks to validate the assumption that an everted tube can be modeled in this way. The tip deflections of everted and uneverted beams under transverse cantilever loads are compared with a tip deflection model that was first developed for aerospace applications. LDPE and silicone coated nylon beams were tested; everted and uneverted beams showed similar tip deflection. The literature model best fit the tip deflection of LDPE tubes with an average tip deflection error of 6 mm, while the nylon tubes had an average tip deflection error of 16.4 mm. Everted beams of both materials buckled at 83\% of the theoretical buckling condition while straight beams collapsed at 109\% of the theoretical buckling condition. The curvature of everted beams was estimated from a tip load and a known displacement showing relative errors of 14.2\% and 17.3\%  for LDPE and nylon beams respectively. This paper shows a numerical method for determining inflated beam deflection. It also provides an iterative method for computing static tip pose and applied wall forces in a known environment.
\end{abstract}

\begin{IEEEkeywords}
\end{IEEEkeywords}


\section{Introduction}
\subsection{Inflated Structures}
Inflated structures have been used for decades in habitats, antennas, wings, and more \cite{Veldman2005}. They are constructed of a skin material that holds load only when inflated. Large structures are commonly a series of smaller tubular components sewn together to create complex shapes \cite{Veldman2005}.  A beam bending model utilized in this paper was developed by Comer \etal for the design of inflatable re-entry vehicles \cite{Comer1963,Fichter1905,Leonard1960}.

\begin{figure} 
\centerline{\includegraphics[width=1\columnwidth]{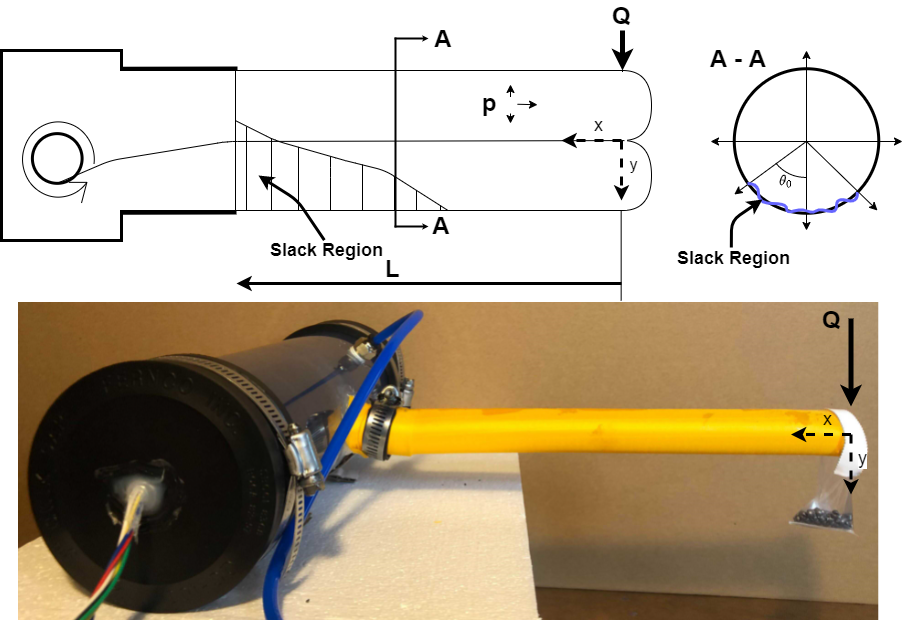}}
\caption{ \textit{Top:} An everted cantilever beam with length, L; internal pressure, p; and external load, Q. Note the coordinate frame is centered at the tip of the beam. \textit{Bottom:} An everted silicone coated nylon beam  carrying a cantilevered tip load. } \label{cantilever_beam}
\end{figure}

When under load, inflated beams exhibit some unique behaviors compared to standard beams. Wrinkles form near the root of the beam, and as the body material wrinkles, it no longer carries tension \cite{Leonard1960,Veldman2005,Comer1963}. This is shown in Fig. \ref{cantilever_beam} as the slack region. As the load increases, the slack region propagates around the beam, and $\theta_0$ increases. As $\theta_0$ approaches $\pi$, the beam will buckle and collapse, behaving like a hinge \cite{Veldman2005,Leonard1960}. 

\subsection{Everted Tubes}
Everted tubes are a subclass of soft robots that grow via tip extension. The robot body is stored within a pressure vessel, usually on a reel, and is deployed as the system is pressurized \cite{Hawkes2017,Blumenschein2020}. Pressure forces at the tip pull the body material outwards, allowing the robot body to grow to arbitrary  lengths. Growth via eversion passively extends the body in the direction of least resistance, allowing it to grow in cluttered and sensitive environments. Everted tubes have been used for exploration, antenna construction, and in medical applications \cite{Hawkes2017,Saxena2020,Slade2017}.

Everted tubes can be constructed from a variety of materials, including thermoplastics (TPU, LDPE), Thermosets, Thermo-coated fabrics, Thermoset-coated fabrics, and uncoated fabrics \cite{Blumenschein2020}. Thermoplastics are the easiest to prototype, as they often can be purchased in manufactured tube shapes of multiple sizes and thicknesses. However, they have the lowest burst pressures and have been shown to fatigue quickly \cite{Blumenschein2020}. Fabricating tubes from thermoset-coated fabrics requires a more involved manufacturing process; they must be sewn and sealed using adhesive. The structure of woven fabric prevents holes from rapidly propagating through the body, preventing bursting, making it ideal for navigating abrasive and rough environments \cite{Blumenschein2020}. 

\subsection{Everted Tube Kinematics and Modeling}
Everted tubes have been modeled as inflated beams in many applications. Many studies have utilized the transverse and axial buckling equations derived in \cite{Comer1963,Fichter1905} to model their behavior. They have also been used to compute smooth retraction forces \cite{Coad2020retraction} and describe kinetic wall interactions \cite{Haggerty2019}, and they have been validated in \cite{Luong2019,Hawkes2017,Greer2017}.

Many novel kinematic capabilities of vine robots have been recently developed, including active steering, smooth retraction, tip-mounted graspers, and dynamically reconfigurable joints   \cite{Blumenschein2020,Coad2020,Hawkes2017,Exarchos2021,Do2020,Greer2017}. These works use external devices along a robot's body to increases its operating workspace and capabilities. However, these rigid additions limit the robot's ability to squeeze through tight spaces since they cannot be crushed and re-inflated, thus hampering some of the benefits of the soft nature of everted tubes.  

Greer \etal provide a differential kinematic model of the free growth of everted tubes. This algorithm accurately describes glancing and head-on growth around obstacles \cite{Greer2018}. These kinematics have been extended by Haggerty \textit{et al.}, who describe the kinetic interactions between an everted tube and wall. The reaction force will cause transverse buckling if the incident tube angle is greater than a theoretical minimum \cite{Haggerty2019}. After a tube buckles from this transverse loading, it can freely grow along the wall while behaving like a hinge \cite{Leonard1960,Haggerty2019}. In an environment with many obstacles, sections of tube between contact points behave as independent beams \cite{Luong2019}. Everted tubes have also been modeled as Cosserat Rods: Selvaggio \etal uses this model to determine the reachable workspace of an everted tube actuated by series Pneumatic Artificial Muscles (sPAM) \cite{Selvaggio2020}. This algorithm utilizes obstacles within the environment to increase the reachability of goal poses. Selvaggio \etal also used the closed-form solution of an externally loaded cantilever Cosserat rod to predict applied environmental forces on such obstacles \cite{Selvaggio2020}. 

In this work, we demonstrate a method for solving the beam model introduced by Comer \etal We use this solution to compare the performance of straight and everted cantilever beams under three different variable loading conditions and validate that everted tubes can be modeled as inflated beams. Current everted tube research uses the maximum axial and transverse loading conditions outlined by \cite{Comer1963,Fichter1905} but have not validated beam deflections or curvatures at loads less than critical loading conditions \cite{Haggerty2019,Luong2019,Greer2017}. Beam deflection was also used to determine an everted tube's curvature, which we have extended to estimate everted tube pose from environmental interactions. Selvaggio \etal estimated this pose using an externally loaded cantilever Cosserat rod for an everted tube with actuation along the length.


\section{Methods}
A rotary eversion device was adapted from designs outlined on \url{vinerobots.org} and by \cite{Hawkes2017}. A 2L pneumatic tank was used to act as a pressure transient filter. Pressure is measured with a gauge pressure transducer (Honeywell, SSCDANN150PG2A3) and filtered using a four frame moving average sampled at 300 Hz. Data was collected on two different eversion devices with pressure transducers calibrated to read within 0.345 kPa, which is within the 2\% error band of the transducer. 

\subsection{Model Derivation}
This study uses the model developed by Comer \etal for an inflated cantilever beam with a length much greater than its radius \cite{Comer1963}. The beam model, as applied to an everted tube, is shown in figure \ref{cantilever_beam}. The curvature 
\begin{equation} \label{curvature_eq}
    \frac{d^{2}y}{dx^2} = 
        \begin{cases}
            \frac{Qx}{EtR^3\pi} & 0<x<\frac{\pi p R^3}{2Q}\\
            \frac{Qx}{EtR^3}\frac{2}{2\pi - 2\theta_0 + sin(2\theta_0)} &  \frac{\pi p R^3}{2Q}<x<L
    \end{cases}
\end{equation}
describes the vertical displacement within the wrinkled and unwrinkled region of the beam. x is the distance along the beam measured from the tip, Q is the applied tip load, E is the Young's Modulus of the material; R is the beam's radius, and t is the material thickness. It is important to note that the origin of the beam is defined at the free end, shown in Fig. \ref{cantilever_beam}, where the load is applied. Displacement downwards is defined in the positive $y$ direction. 

This curvature equation is similar to the standard cantilever beam equation of $\kappa = \frac{M}{EI}$, where the inertia of a thin-walled cylinder is $I=\pi R^3t$. $\theta_0$ is the wrinkle angle around the beam, shown in figure \ref{cantilever_beam}. Wrinkle angle is numerically approximated from eq. \ref{wrinkle_angle_eq} using a  5\textsuperscript{th} order polynomial and is described as function of $\frac{Qx}{pR^3}$, eq. \ref{th0_fn}. This process is outlined by \cite{Veldman2005}. 

\begin{equation}\label{wrinkle_angle_eq}
    \frac{Qx}{pR^3} = \frac{\pi ( 2\pi - 2\theta_{0} + sin(2\theta_{0})}{4[sin(\theta_0) + (\pi - \theta_0)cos(\theta_0)]}
\end{equation}

The problem is simplified by non-dimensionalizing the position in the x and y direction. The following substitutions:
\begin{equation}\label{xi}
    \xi = (\frac{Q}{p R^3})x,
\end{equation}
    
\begin{equation}\label{eta}
    \eta = (\frac{Q^2 Et}{p^3 R^3})y.
\end{equation}

\begin{equation} \label{th0_fn}
    \theta_0 = f(\frac{Qx}{p R^3}) = f(\xi)
\end{equation}

give a non-dimensional model for curvature:

\begin{equation} \label{curvature_nd}
    \frac{d^2\eta}{d\xi^2} = 
    \begin{cases}
    \frac{\xi}{\pi} & \frac{\pi}{2}>\xi>0\\
    \xi \frac{2}{2\pi - 2\theta_0 + sin(\theta_0)} & \pi>\xi>\frac{\pi}{2}
    \end{cases}.
\end{equation}

We solve the beam by decomposing the non-dimensional model into a system of first order equations by $\eta_1$ and $\eta_2$, where $ \eta_1 = \eta,  \; \eta_2=\frac{d\eta}{d\xi}$:

\begin{equation} \label{nd_sys}
    \frac{d}{d\xi} \begin{bmatrix} \eta_1 \\ \eta_2 
    \end{bmatrix} =
    \begin{bmatrix}
    \eta_2 \\  \begin{cases}
    \xi \frac{2}{2\pi - 2\theta_0 + sin(\theta_0)}
    & \pi>\xi>\frac{\pi}{2}\\
    \frac{\xi}{\pi} & \frac{\pi}{2}>\xi>0
    \end{cases}
    \end{bmatrix}.
\end{equation}


Using the initial conditions described in \cite{Comer1963}, a solution can be found. The initial conditions are that the displacement and slope at the root ($x=L$) of the beam are both zero. The system is solved by numerically solving the initial value problem backward, simulating from $x=L$ to $x=0$. Matlab's ODE solver function, ode45, was used to solve

$$ 
\eta(\xi(L)) = 0, \;\;\;
\frac{d\eta(\xi(L))}{d\xi} = 0
$$
$$ \xi(L) = \frac{QL}{pR^3}.
$$

\subsection{Mechanics of Materials}
Two different beam materials were used to validate the inflated beam model: Silicone Coated Nylon (Seattle Fabrics, Seattle, WA, USA) and 2.54 cm diameter with 2mil (0.05mm) wall thickness Low-Density-PolyEthylene (LDPE) tubing (ULINE, Pleasant Prairie, WI, USA).
The Silicone Coated Nylon had a measured thickness of 0.12 mm. Beams were sewn to a 2.54 cm diameter and sealed using Seam Grip WP (Gear Aid, Bellingham, WA, USA). The LDPE tubing held a significant amount of memory from its manufacturing and storage on a large reel, giving all inflated tubes a nominal curvature. To eliminate the curvature in the plastic was annealed by hanging vertically and stretched by weights in a sunny enclosed patio. 

Each material was tensile tested, the stress and strain were computed using a video extensometer and digital image correlation. Materials were stretched at a rate of 5mm/min. Both annealed and un-annealed LDPE samples were tested to ensure that the annealing process did not affect the mechanical properties of the plastic. The Young’s modulus was measured between the maximum and minimum stress within the beam during the following conditions: length of 0.357 m, internal pressure of 10.34 kPa, and applied tip load of 0.155 N. The minimum stress is the longitudinal stress of a pressure vessel and the maximum axial stress at the root of the loaded cantilever beam ($x=L$), outlined in \cite{Comer1963}:

\begin{equation} \label{max_stress}
    \sigma_m = \frac{QL}{tR^2}\frac{2(1+cos(\theta_0))}{2\pi-2\theta_0+sin(2\theta_0)}.
\end{equation}

\subsection{Parameter Variation}
The model was validated by comparing the tip deflection of straight and everted tubes. Straight tubes are defined as traditional cylindrical thin-walled inflated beam. In contrast everted tubes have an inner lumen, or tail, connecting back to the spindle, shown in Fig. \ref{cantilever_beam}. Tests varied the independent values of eq. \ref{eta}: beam length (L), internal pressure (p), and external load (Q). Silicone coated nylon and annealed LDPE were the two tube materials used. The tip displacement is $y_d = \eta \frac{p^3 R^6}{Q^2 E t}$ for $\xi(0)$. Transverse buckling occurred when the beam collapsed to the floor under the test conditions. 

For beams under variable tip load, a 0.368 m beam with a 1.27 cm radius and 10.34 kPa internal pressure was loaded at the tip by weights of increasing mass. The tip displacement was measured between every increase in weight, the weight was also removed between increases. The beam was loaded until it experienced transverse buckling and collapsed. Collapse conditions were compared to the theoretical critical values \cite{Comer1963}:

\begin{equation} \label{collapse_conditions}
Q_{max} = \frac{\pi p R^3}{L}
\end{equation}

A new plastic beam was used after each collapse to avoid fatigue and plastic deformation between trials. Because the nylon had a significantly higher stiffness, beams were depressurized and re-oriented after every collapse to mitigate any seam dependent deflection. 

Everted and straight beams of variable length were evaluated by pressurizing to 10.34 kPa and tip loading with a 0.155 N, tip deflection was measured. This process was repeated for beams of increasing length until the beam collapsed under load. Beams were similarly replaced or re-oriented after each collapse. Collapse conditions were compared to the theoretical max length. 

Beams of varying internal pressure were also evaluated using a constant length of 0.357 m and mass of 0.155 N. The tube's internal pressure was decreased from 27.58 kPa until collapse occurred. All beams were depressurized and unloaded between each trial. 

 Eq. \ref{collapse_conditions} was used to determine additional critical buckling conditions, max length ($L_{max}$) and minimum internal pressure ($P_{min}$). 

\subsection{Curvature Evaluation}
\begin{figure}[t!]
\centerline{\includegraphics[width=1\columnwidth]{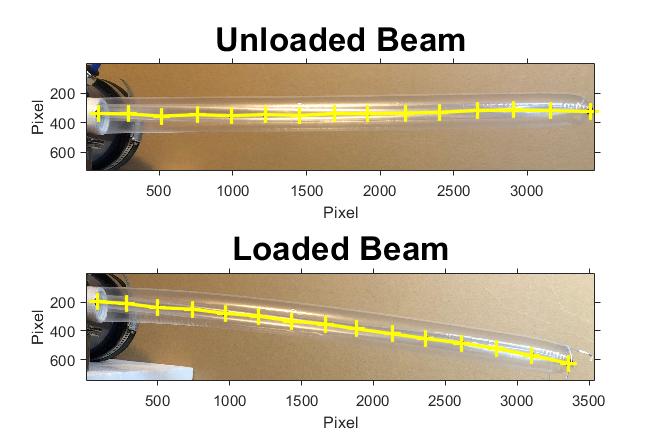}}
\caption{Everted inflated cantilever beam with and without a tip load. An image mask was used to determine the centroid of each marker to determine relative displacements along the beam.}
\label{loaded_beam}
\end{figure}

Annealed LDPE everted beams of variable length, and constant pressure of 10.34 kPa were loaded with a 0.155 N tip load. Each beam was marked with a series of black dots at 2.54 cm increments. Images were taken before and during loading. Their relative vertical displacement was measured using an image mask in Matlab (Fig. \ref{loaded_beam}). Displacement at discrete locations along the beam was compared with the inflated beam model. The tip slope was determined from the two distal-most markers and compared with the modeled beam slope, $\eta_2$, at the tip. 

\subsection{Curvature Evaluation From Environmental Interactions}
The model was used to predict tube curvature from environmental interactions. An annealed LDPE everted tube was tip-displaced a fixed amount. Given constant parameters of length and internal pressure, the theoretical load at the tip was iteratively determined, and the curvature estimated from the model. A tube with tip displacement greater than the theoretical maximum was modeled as buckled, and the curvature was approximated as a straight line between tip and base, as presented in \cite{Luong2019}. 


\section{Results}

\subsection{Materials Testing}
Annealed and non-annealed LDPE samples were loaded to between 10\% and 12\% strain. The Young's Modulus was determined from the elastic region, $<2\%$ strain, to be 199 MPa and 243 MPa for the non-annealed and annealed samples, respectively. All subsequent simulations use the average across all LDPE conditions, 227 MPa. The Young's Modulus of silicone coated nylon was linear within the max/min stress region, consistent with \cite{Leonard1960}, and measured to be 495 MPa.

\begin{figure}[t!]
\centerline{\includegraphics[width=1\columnwidth]{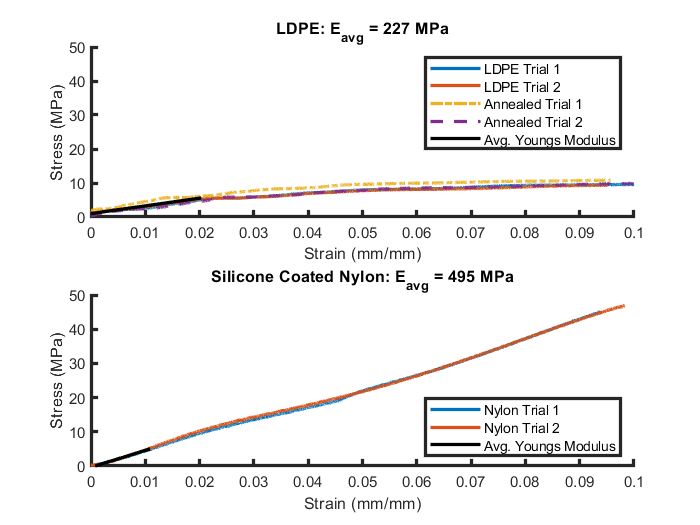}}
\caption{Stress-strain performance of the tested materials. \textit{Top:} LDPE results, a comparison between the annealed and unannealed material shows approximately the same Young’s modulus of 227MPa. \textit{Bottom:} Silicone coated nylon was strained up to 10\%, the data shows that this is still in the elastic region. The Young’s Modulus was calculated within experimental stress values. 
}
\label{instron_results}
\end{figure}

\subsection{Model Validation: Variable Tip Load} 
\begin{figure}[t!]
\centerline{\includegraphics[width=1\columnwidth]{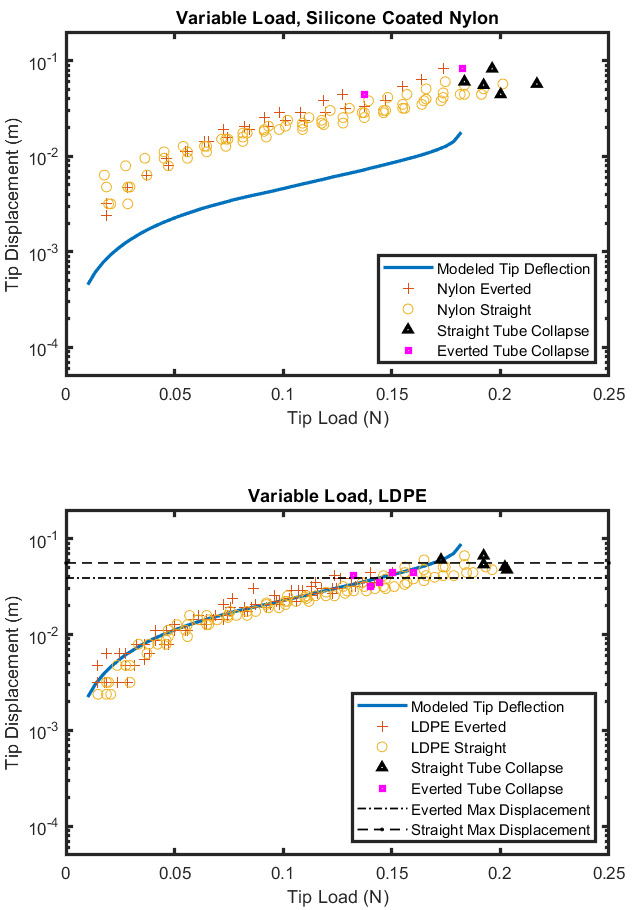}}
\caption{Tip deflection of straight and everted inflated beams under variable load at a constant length of 0.357 m and constant pressure of 10.34 kPa. The $\square$ and $\triangle$ markers represent the load at which the beam collapsed.  \textit{Top:} Silicone coated nylon beam. 
\textit{Bottom:} Annealed LDPE beam. 
}
\label{var_load}
\end{figure}

A constant length, constant pressure, cantilever beam under variable tip load was tested. Fig. \ref{var_load} shows measured tip deflection against applied tip load. A solid blue line shows the theoretical tip deflection until the theoretical buckling condition and displacement. Absolute error was calculated as the difference between the model and measured tip deflection at a specific load. Beam collapse conditions are shown with $\square$ and $\triangle$ markers and evaluated as a percentage of the theoretical maximum load condition. Buckling displacement is the measured tip deflection at the loading condition just before the experimental buckling condition. Everted and straight nylon beams had average absolute tip deflection errors of 19.3 and 16 mm, respectively. The everted beams collapsed at 88\% of the theoretical maximum load, while the straight tubes collapsed at 107\% of the theoretical maximum. The LDPE tubes had a lower absolute tip deflection error of 2.2 and 3.8 mm for everted and straight tubes, respectively. On average, everted LDPE beams collapsed at 80\% of the theoretical maximum load at an average displacement of 0.039 m (SD: 7 mm). Straight LDPE beams collapsed at 106\% of the theoretical maximum load at an average displacement of 0.056 m (SD: 5 mm). The theoretical maximum displacement at $Q_{max}$ is 0.088 m. 

\subsection{Model Validation: Variable Beam Length}
\begin{figure}[t!]
\centerline{\includegraphics[width=.925\columnwidth]{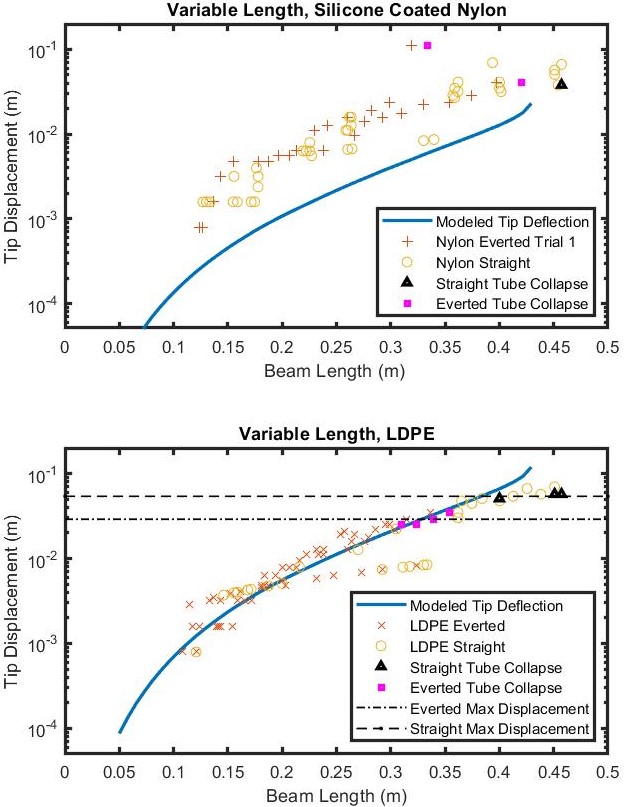}}
\caption{Tip deflection of straight and everted beams under constant 15.8 g load, pressurized to 10.34 kPa under variable lengths.The $\square$ and $\triangle$ markers represent the load at which the beam collapsed. \textit{Top:} Silicone coated nylon beam. 
\textit{Bottom:} Annealed LDPE beam. 
}
\label{var_len}
\end{figure}

A constant pressure, constant load cantilever beam with variable length was tested (Fig. \ref{var_len}).  A solid blue line shows the theoretical tip deflection until the theoretical buckling length and displacement. Absolute error was calculated as the difference between the model and measured tip deflection at a specific beam lengths. Collapse conditions are shown with $\square$ and $\triangle$ markers and evaluated as a percentage of the theoretical maximum length. Buckling displacement is the measured tip deflection at the beam length just before the experimental buckling length. Everted and straight nylon beams had absolute tip deflection errors of 12.9 and 8.3 mm, respectively. The everted beams collapsed at 88\% of the theoretical maximum load, while the straight tubes collapsed at 106\% of the theoretical maximum. The LDPE tubes had lower absolute tip deflection errors of 2.5 and 8.6 mm for everted and straight, respectively. Everted LDPE beams collapsed at 77\% of the theoretical maximum load at an average displacement of 0.029 m (SD: 4 mm). Straight LDPE beams collapsed at 102\% of the theoretical maximum load at an average displacement of 0.054 m (SD: 3 mm). The theoretical maximum displacement at $L_{max}$ is 0.119 m. 

\subsection{Model Validation: Variable Pressure}
\begin{figure}[t!]
\centerline{\includegraphics[width=.925\columnwidth]{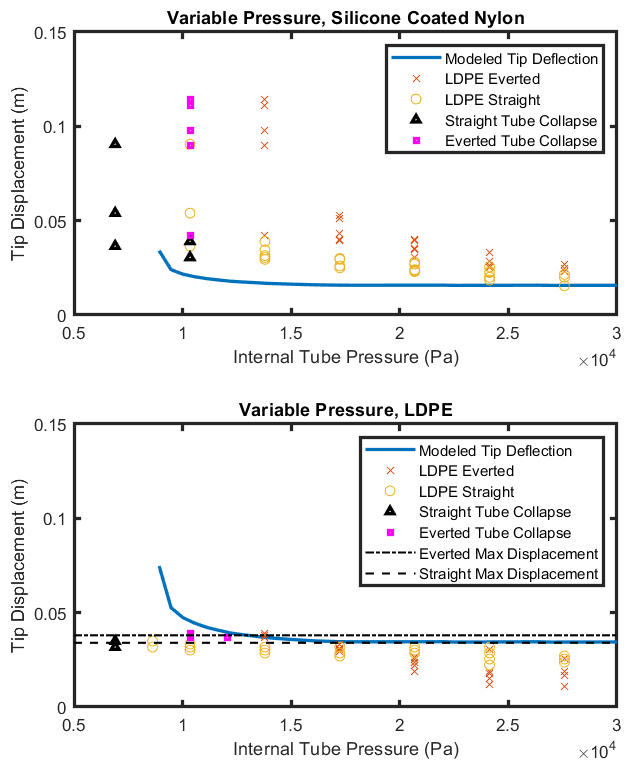}}
\caption{Tip deflection of straight and everted beams of variable pressure under constant 0.155 N load and length of 0.357 m. The $\square$ and $\triangle$ markers represent the load at which the beam collapsed. \textit{Top:} Silicone coated nylon beam.
\textit{Bottom:} Annealed LDPE beam. 
}
\label{var_pres}
\end{figure}

A constant load, constant length cantilever beam with variable internal pressure was tested. The results are shown in Fig. \ref{var_pres} which plots tip deflection against internal pressure. A solid blue line shows the theoretical tip deflection from the theoretical minimum internal pressure to the deflection at 30kPa. Absolute error was calculated as the difference between the modeled and measured tip deflection at a specific internal pressure. Collapse conditions are shown with $\square$ and $\triangle$ markers and evaluated as a percentage of the theoretical minimum pressure. Buckling displacement is the measured tip deflection at the internal pressure just before the experimental buckling pressure. Everted and straight nylon beams had an absolute tip deflection error of 29.1 and 12.9 mm, respectively. Nylon everted beams, on average, collapsed at 116\% of the theoretical minimum pressure while the straight beams collapsed at 93\% of the theoretical minimum. LDPE tubes had a lower absolute tip deflection of error of 9.3 and 8.2 mm for everted and straight beams, respectively. The everted LDPE beams collapsed at 120\% of the theoretical minimum pressure at a displacement of 0.038m (SD: 1 mm). Straight beams collapsed at 77\% of the theoretical minimum at a displacement of 0.034m (SD: 1 mm). The theoretical maximum displacement at ($P_{min})$ is 0.0747m. 

\begin{figure*}[t!]
\centerline{\includegraphics[width=0.99\textwidth]{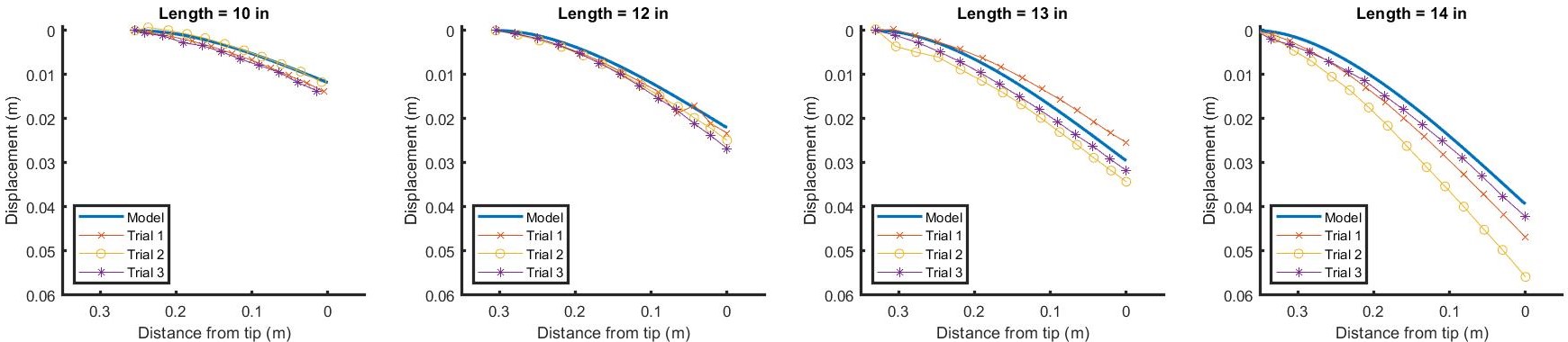}}
\caption{Modeled and measured curvature for everted LDPE beams of length 0.254, 0.305, 0.330, and 0.356 m. Each beam was under a 0.155 N load and pressurized to 10.34 kPa. }
\label{curvature}
\end{figure*}

\subsection{Model Validation: Curvature Under Load}
The curvature of an everted cantilever beam of variable length is shown in Fig. \ref{curvature}. The origin is located at the tip of the beam, and the plot shows distance from the tip against relative displacement. The absolute displacement error was computed as the difference between the measured vertical displacement and modeled vertical displacement at discrete locations along the length of the beam. The average displacement error was computed to be  1.1, 1.5, 2.6 and 4.9 mm for beams of length 0.254, 0.305, 0.330, and 0.356m. While the absolute error increases with beam length, the relative tip displacement error is 14\%, 11.8\%, 12.4\%, and 18.5\%. The tip slope error, $\frac{dy}{dx}$, was measured to be 0.011, 0.017, 0.012, and 0.02 rad for the given lengths.

\subsection{Estimating Curvature from Environmental Interaction}
\begin{figure*}[t!]
\centerline{\includegraphics[width=1\textwidth]{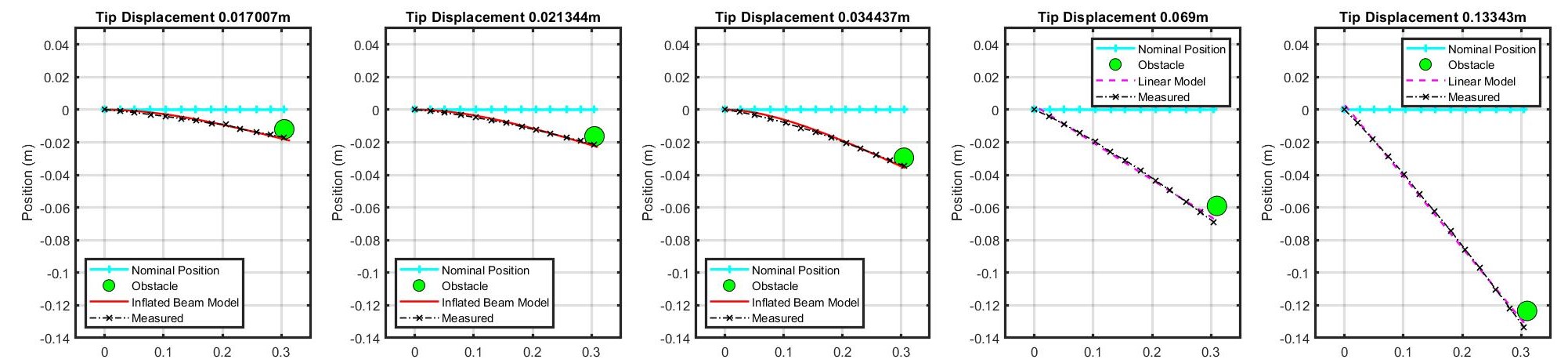}}
\caption{A comparison of the modeled and measured curvature for beams deflected by an obstacle in the environment. A beam 0.305 m in length and pressurized to 10.34 kPa was deflected to a known distance. Its curvature was measured using a series of markers spaced 2.54 cm apart. Given the tip displacement, length, and pressure, the applied force and curvature were estimated.}
\label{obstacle}
\end{figure*}

An inflated beam's curvature was estimated from a known tip displacement by iteratively solving for the theoretical applied load, given constant beam parameters. Absolute error was calculated from the difference between the estimated curvature and the measured curvature. Figure \ref{obstacle} shows the beams interaction with an obstacle relative to its nominal unloaded position. For tip displacements of 17, 21, and 34 mm an absolute error of 0.7, 0.7, and 0.9 mm and relative error of 20\%, 17\% and 15\% was calculated, respectively. The curvature of beams displaced greater than their theoretical max, determined using eq. \ref{collapse_conditions}, were modeled as a straight line between the tip and base. This straight-line approximation of buckled beams recorded average absolute errors of 1.1 and 1.4 mm and relative error of 4.5\% and 3\%. 

\section{Discussion}
\subsection{Tip Deflection Validation}
The deflection model provided by Comer \etal best fit the tip deflections of the annealed LDPE tubes with an average tip deflection error of 6 mm for both straight and everted beams. The measured tip deflection of silicone coated nylon did not match the magnitude predicted in the model with an average tip deflection error of 16.4 mm for both straight and everted beams. The model held the same shape as the experimental tip deflection data. In this cantilever application, modeled nylon tubes exhibited an effective stiffness of approximately 25\% of the measured value. The data shows that $E_{model} = .25E_{longitudinal}$, where $E_{longitudinal}$ is the Young's Modulus computed from the tensile test. This is likely due to the direction of the applied load on the weave pattern of the nylon. Silicone coated nylon can be thought of as a composite material whose Young's Modulus is not constant with respect to the angle of the applied load. Tensile testing yielded the stiffness in the longitudinal direction. Cantilever loading applies transverse stress on the beam, a loading condition where $E_{transverse}<E_{longitudinal}$. For a structure where the load may be applied in an arbitrary direction with respect to the weave, performance consistent with the maximum material properties determined from longitudinal testing cannot be expected. 

In all cases, the everted and straight tubes showed approximately the same tip deflection throughout each experiment. Although in all experiments, everted tubes collapsed at conditions before the theoretical max/min and straight tubes collapsed after. Across all experiments, everted nylon tubes buckled 86\% of the theoretical buckling condition while everted LDPE tubes collapsed at 79\% of the theoretical buckling condition. On average, straight nylon tubes buckled at 107\% of the theoretical buckling condition and straight LDPE tubes collapsed at 110\% of the theoretical buckling condition. This could be due to wrinkles or fatigue generated in the material during the eversion process. Recall that as $\lim_{\theta_0\to\pi}$ the beam collapses \cite{Veldman2005}. It is possible that material creasing during eversion contributed wrinkle propagation when under load. The deflection of an inflated cantilever beam model is valid for homogeneous materials such as LDPE. The modeled beam's deflection held true until near collapse, but the collapse condition may not be accurately determined by the model. A beam's collapse may be defined better by a maximum displacement. Across all LDPE tests, beams collapsed at a consistent displacement rather than load. In nearly all conditions, everted tubes collapsed at a smaller displacement than straight tubes. 

\subsection{Beam Curvature Validation}
The curvature of an everted tube was validated using both a transverse gravity load and external lateral displacement. Tubes under a transverse gravity load had a relative tip displacement error of 14.2\%, and tubes under an external lateral displacement had a relative curvature error of 17.3\%. Tubes externally displaced greater than the theoretical maximum were approximated as a straight line and showed an error of 3.75\%. Longer tubes showed greater absolute tip displacement error but all beams had similar relative tip error. Tip slope computation showed a similar error trend across beams of variable length. The accuracy of this value can be increased by increasing the quality of the image tracking system. 
Predicting curvature from environmental displacements implies that the model may be used to accurately estimate interaction forces with the wall given a specific tip deflection. Allowing a user to determine applied forces in a known sensitive environment such as an archaeological dig site, \cite{Coad2020}, or within the body \cite{Slade2017, Saxena2020}. 

\begin{figure}[h]
\centerline{\includegraphics[width=.75\columnwidth]{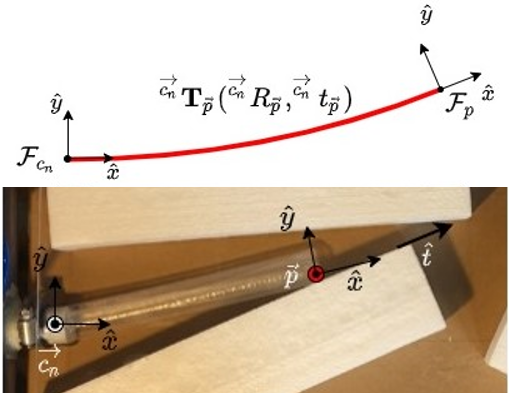}}
\caption{Tip pose estimation from static tip deflection}
\label{pose_estimation}
\end{figure}

\subsection{Pose Estimation from Environment}
The static pose of the tip of an everted tube can be estimated using tip deflection. The tip reference frame is calculated using the differential kinematics model derived by \cite{Greer2018}:
\begin{equation} \label{p_dot_vec}
\dot{\vec{p}} = u\frac{||\vec{p}-\vec{c_n}||}{\hat{t}\cdot(\vec{p}-\vec{c_n})}
\end{equation}
Given an obstacle within an environment, tip translation is determined by :
\begin{equation} \label{trans_l}
     ^{\vec{c_n}}\mathbf{t}_{\vec{p}} = \vec{p},
\end{equation}
where $\hat{t}$ is the unit vector parallel to the surface of the obstacle, $u$ is the eversion growth rate, $\hat{p}$ is the tip location, and $c_n$ is the nearest contact point. In the case of Fig. \ref{pose_estimation} $c_n$ is the everter base. The curvature at the tip or heading, $\eta_2$ when $\xi=0$, gives the rotation angle of the tip coordinate frame:
\begin{equation} \label{rot_vec}
     ^{\vec{c_n}}\mathbf{R}_{\vec{p}} = \eta_2 \frac{p^2 R^3}{QEt}
\end{equation}

From tip translation and tip slope, a planar and homogeneous transformation matrix can be determined. This process is similarly computed in Selvaggio \etal \cite{Selvaggio2020}, where the everted tube is modeled as a Cosserat rod. Forces applied to the obstacle can also be computed from the pose.

\subsection{Applications}
Understanding the static behavior of everted beams will help give insight into their kinematic behavior. This work can support the algorithms derived in several published works. 

Tip slope angle could also be useful in determining transverse buckling while growing along a wall. If a tube is incident to a wall at an angle greater than the minimum incident angle described by Haggerty \etal \cite{Haggerty2019}, transverse buckling occurs and the tube grows along the wall. If a tube is incident to the wall at an angle less than the determined minimum incident angle, the static beam will bend according to the model described by \cite{Comer1963} and will behave as shown in Fig. \ref{obstacle}. In this instance, tip slope angle should be considered when computing incident angle. Transverse buckling may occur when tip slope is considered. This insight could help understand a tube's kinematic behavior very close to buckling conditions.

Retraction without buckling of an everted tube is greatly influenced by the tube's curvature \cite{Coad2020retraction}. The curvature model derived in this paper can provide an analytical solution to assist in computing retraction forces. 

Everted tubes in previous path planning algorithms have been modeled as straight lines between obstacles and discrete pivot points \cite{Greer2018}. The inflated beam model allows for the beam's curvature to be computed and included in these kinematic models.

\subsection{Limitations and Future Work}
While the solution of an inflated cantilever beam is easy to compute and useful in static conditions, it does not have a closed-form solution. Determining theoretical max tip deflection or applied wall force from tip deflection requires iterative solving, which may be too slow for real-time control. 

In the future, bi-axial testing of composite tube materials should be conducted to improve models of transverse buckling. Additionally, experiments should be done to extend these models to better describe the kinematics of everted tubes. Active pressure control should be studied for the kinematic control of everted tubes, thus maintaining the benefits of soft robots.

\section{Conclusion}
This study validates the current assumption that an everted tube can be modeled like an inflated beam. Our results confirm that the deflection and curvature of an LDPE everted tube can be accurately modeled using inflated beam theory. Beams composed of composite materials, like silicone-coated nylon, are not as accurately modeled. However, the model overestimates the exact buckling condition across all materials tested. Data indicates that everted tube buckling was better described by a maximum displacement rather than a maximum loading condition. The model can be iteratively solved to determine curvature from environmental displacements and applied loads to the environment. 

\section{Acknowledgements}
This work was supported by the UW Burke Center for Entrepreneurship's Prototype Fund and UW CoMotion's Innovation Gap Fund. Thanks Mark Gerges and the UW MACS lab for 3D printing assitance.

\bibliographystyle{IEEEtran}
\bibliography{references.bib}
\end{document}